*Research Note*

# Characterizations of Decomposable Dependency Models

**Luis M. de Campos** LCI@DECSAI.UGR.ES
*Departamento de Ciencias de la Computación e I.A.*
*E.T.S. Ingeniería Informática, Universidad de Granada*
*18071 - Granada SPAIN*

## Abstract

Decomposable dependency models possess a number of interesting and useful properties. This paper presents new characterizations of decomposable models in terms of independence relationships, which are obtained by adding a single axiom to the well-known set characterizing dependency models that are isomorphic to undirected graphs. We also briefly discuss a potential application of our results to the problem of learning graphical models from data.

## 1. Introduction

Graphical models are knowledge representation tools commonly used by an increasing number of researchers, particularly from the Artificial Intelligence and Statistics communities. The reason for the success of graphical models is their capacity to represent and handle independence relationships, which have proved crucial for the efficient management and storage of information (Pearl, 1988).

There are different kinds of graphical models, although we are particularly interested in undirected and directed graphs (which, in a probabilistic context, are usually called Markov networks and Bayesian networks, respectively). Each one has its own merits and shortcomings, but neither of these two representations has more expressive power than the other: there are independence relationships that can be represented by means of directed graphs (using the d-separation criterion) and cannot be represented by using undirected ones (through the separation criterion), and reciprocally. However, there is a class of models that can be represented by means of both directed and undirected graphs, which is precisely the class of decomposable models (Haberman, 1974; Pearl, 1988). Decomposable models also possess important properties, relative to factorization and parameter estimation, which make them quite useful. So, these models have been studied and characterized in many different ways (Beeri, Fagin, Maier, & Yannakakis, 1983; Haberman, 1974; Lauritzen, Speed, & Vijayan, 1984; Pearl, 1988; Wermuth & Lauritzen, 1983; Whittaker, 1991). For example, decomposable models have been characterized as the kind of dependency models isomorphic to chordal graphs (Lauritzen et al., 1984; Whittaker, 1991).

However, we do not know any characterization of decomposable models in terms of the kind of independence relationships that they are capable of representing. This is somewhat surprising, because it seems quite natural to us to characterize a type of object using the same terms as those used to define it; in our case, the object is a special type of dependency model, i.e., a collection of conditional independence statements about a set of variables in a given domain of knowledge, and therefore we should be able to describe it in terms of properties of these independence relationships. The objective of this paper is precisely to obtain such a characterization of decomposable models.

Our approach to the problem will be based on identifying the set of properties or axioms that a collection of independence relationships must satisfy, in order to be representable by a chordal graph. This approach has been successfully used to study other kinds of dependency models: Pearl and Paz (1985) identified the set of properties characterizing models isomorphic to undirected graphs, and de Campos (1996) determined the axioms that characterize models isomorphic to undirected and directed singly connected graphs (i.e., trees and polytrees, respectively).

The rest of the paper is organized as follows. In Section 2 we briefly describe several concepts which are basic for subsequent development. Section 3 introduces decomposable models and their representation using chordal graphs. In Section 4 we prove two characterizations of decomposable models. These characterizations turn out to be surprisingly simple: we only have to add a single property to the set of axioms characterizing dependency models isomorphic to undirected graphs. Section 5 discusses the relationships between our results and Lauritzen's characterization of chordal graphs. Finally, Section 6 contains the concluding remarks and some proposals for future work, which include the application of the results developed here to the problem of learning graphical models from data.

## 2. Preliminaries

In this section, we are going to describe the notation as well as some basic concepts and results used throughout the paper.

A *Dependency Model* (Pearl, 1988) is a pair $M = (U, I)$, where $U$ is a finite set of elements or variables, and $I(.,.|.)$ is a rule that assigns truth values to a three place predicate whose arguments are disjoint subsets of $U$. Single elements of $U$ will be denoted by standard or Greek lowercase letters, whereas subsets of $U$ will be represented by capital letters. The interpretation of the conditional independence assertion $I(X, Y|Z)$ is that having observed $Z$, no additional information about $X$ could be obtained by also observing $Y$. For example, in a probabilistic model (Dawid, 1979; Lauritzen, Dawid, Larsen, & Leimer, 1990), $I(X, Y|Z)$ holds if and only if

$$P(\mathtt{x}|\mathtt{z}, \mathtt{y}) = P(\mathtt{x}|\mathtt{z}) \text{ whenever } P(\mathtt{z}, \mathtt{y}) > 0,$$

for every instantiation $\mathtt{x}$, $\mathtt{y}$ and $\mathtt{z}$ of the sets of variables $X$, $Y$ and $Z$. However, dependency models are applicable to many situations far beyond probabilistic models (de Campos, 1995; Pearl, 1988; Shenoy, 1992).

A graphical representation of a dependency model $M = (U, I)$ is a direct correspondence between the elements in $U$ and the set of nodes in a given graph, $G$, such that the topology of $G$ reflects some properties of $I$. The topological property selected to represent independence assertions depends on the type of graph we use: *separation* for undirected graphs and *d-separation* (Pearl, 1988; Verma & Pearl, 1990) for directed acyclic graphs (dags):

- *Separation*: Given an undirected graph $G$, two subsets of nodes, $X$ and $Y$, are said to be separated by the set of nodes $Z$, and this is denoted by $\langle X, Y|Z\rangle_G$, if $Z$ intercepts all chains between the nodes in $X$ and those in $Y$.

- *d-separation*: Given a dag $G$, a chain $C$ (a chain in a directed graph is a sequence of adjacent nodes, the direction of the arrows does not matter) from node $\alpha$ to node $\beta$ is said to be blocked by the set of nodes $Z$, if there is a vertex $\gamma \in C$ such that, either
  - $\gamma \in Z$ and arrows of $C$ do not meet head to head at $\gamma$, or
  - $\gamma \notin Z$, nor has $\gamma$ any descendants in $Z$, and the arrows of $C$ do meet head to head at $\gamma$.

  Two subsets of nodes, $X$ and $Y$, are said to be d-separated by $Z$, and this is also denoted by $\langle X, Y|Z\rangle_G$, if all chains between the nodes in $X$ and the nodes in $Y$ are blocked by $Z$. There exists a criterion equivalent to d-separation, based on the separation of $X$ from $Y$ by $Z$ in the moral graph of the smallest ancestral set containing $X \cup Y \cup Z$ (Lauritzen et al., 1990).

Given a dependency model, $M$, we say that an undirected graph (a dag, respectively), $G$, is an *I-map* if every separation (d-separation, respectively) in $G$ implies an independence in $M$: $\langle X, Y|Z\rangle_G \Rightarrow I(X, Y|Z)$. On the other hand, an undirected graph (a dag, resp.), $G$, is called a *D-map* if every independence relation in the model implies a separation (d-separation resp.) in the graph: $I(X, Y|Z) \Rightarrow \langle X, Y|Z\rangle_G$. A graph, $G$, is a *Perfect map* of $M$ if it is both an I-map and a D-map. $M$ is said to be *graph-isomorphic* if a graph exists which is a perfect map of $M$.

The class of dependency models isomorphic to undirected graphs has been completely characterized (Pearl & Paz, 1985) in terms of five properties or axioms satisfied by the independence relationships within the model:

(C1) Symmetry:
$(I(X, Y|Z) \Rightarrow I(Y, X|Z)) \, \forall X, Y, Z \subseteq U$.

(C2) Decomposition:
$(I(X, Y \cup W|Z) \Rightarrow I(X, Y|Z)) \, \forall X, Y, W, Z \subseteq U$.

(C3) Strong Union:
$(I(X, Y|Z) \Rightarrow I(X, Y|Z \cup W)) \, \forall X, Y, W, Z \subseteq U$.

(C4) Intersection:
$(I(X, Y|Z \cup W)$ and $I(X, W|Z \cup Y) \Rightarrow I(X, Y \cup W|Z)) \, \forall X, Y, W, Z \subseteq U$.

(C5) Transitivity:
$(I(X, Y|Z) \Rightarrow I(X, \gamma|Z)$ or $I(\gamma, Y|Z) \, \forall \gamma \in U \setminus (X \cup Y \cup Z)) \, \forall X, Y, Z \subseteq U$.

Pearl and Paz also tacitly assumed that an additional, trivial, axiom holds, namely $I(X, \emptyset|Z)$ $\forall X, Z \subseteq U$. They also assumed all through that the sets $X, Y, Z, W$ involved in the axioms are pairwise disjoint.

**Theorem 1 (Pearl and Paz, 1985)** *A dependency model $M$ is isomorphic to an undirected graph if, and only if, it satisfies the axioms C1–C5.*

The graph associated with the dependency model $M$, such that conditional independence in $M$ is equivalent to separation in this graph, is $G_M = (U, E_M)$, where the set of edges

$E_M$ is

$$E_M = \{\alpha\text{–}\beta \mid \alpha, \beta \in U,\ \neg I(\alpha, \beta | U \setminus \{\alpha, \beta\})\}.$$

On the other hand, the class of dependency models isomorphic to dags is considerably more difficult to characterize. It has been suggested (Geiger, 1987; Pearl, 1988) that the number of axioms required for a complete characterization of the d-separation in dags is probably unbounded. However, some more restricted models, namely polytree-isomorphic models, can be fully characterized by using a finite number of axioms (de Campos, 1996).

Graphical models are not only convenient means of expressing conditional independence statements in a given domain of knowledge, they also convey information necessary for decisions and inference, in the form of numerical parameters quantifying the strength of each link. The assignment of numerical parameters to a graphical model is also quite different for undirected and directed graphs (here we restrict the discussion to probabilistic models). In the case of directed acyclic graphs, this is a simple matter: we only have to assign to each variable $x_i$ in the dag a conditional probability distribution for every instantiation of the variables that form the parent set of $x_i$, $\pi(x_i)$. The product of these local distributions constitutes a complete and consistent specification, i.e., a joint probability distribution (which also preserves the independence relationships displayed by the dag):

$$P(x_1, x_2, \ldots, x_n) = \prod_{i=1}^{n} P(x_i | \pi(x_i))$$

However, the case of undirected graphs is different: constructing a complete and consistent quantitative specification while preserving the dependence structure of an arbitrary undirected graph can be done using the method of Gibb's potentials (Lauritzen, 1982) (which assigns compatibility functions to the cliques of the graph), but it is considerably more complicated, in terms of both computational effort and meaningfulness of the parameters, than the simple method used for dags.

## 3. Decomposable Models and Chordal Graphs

Some dependency models representable by means of a special class of undirected graphs do not present the quantification problem described above. These are the so called decomposable models, which also exhibit a number of important and useful additional properties. There are several ways of defining decomposable models. The most appropriate to our interests, which mainly lie in graphical modelling, is based on a graph-theoretic concept: chordal graphs, also called triangulated graphs (Rose, 1970).

**Definition 1** *An undirected graph is said to be chordal if every cycle of length four or more has a chord, i.e., an edge linking two non-adjacent nodes in the cycle.*

The simplest example of a non-chordal graph is the diamond-shaped graph displayed in Figure 1 (a).

**Definition 2** *A dependency model is decomposable if it is isomorphic to a chordal graph.*

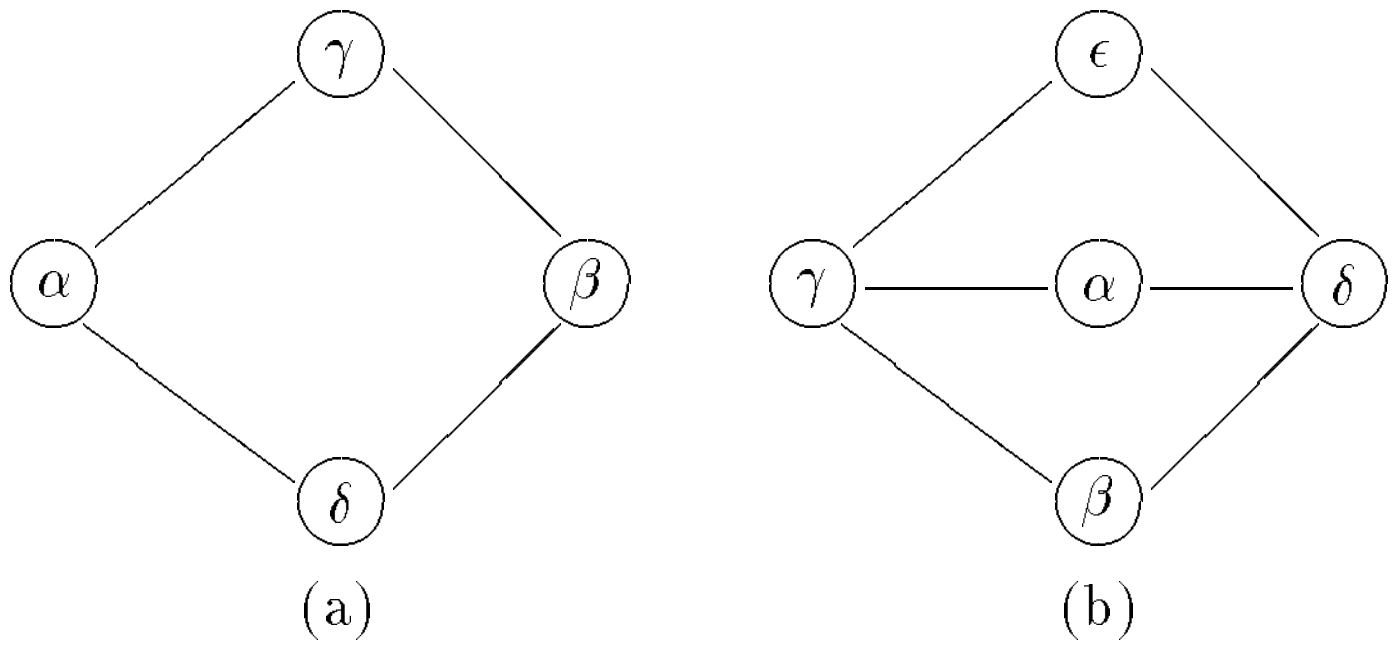

Figure 1: (a) The simplest example of a non-chordal graph (b) Non-chordal graph satisfiying C1–C5 and C7

One important property satisfied by every chordal graph $G$, which in fact characterizes chordal graphs (Beeri et al., 1983), is that the edges of $G$ can be directed acyclically so that every pair of converging arrows emanates from two adjacent nodes. From this property, it can be deduced (Pearl, 1988) that the class of dependency models that may be represented by both a dag and an undirected graph is precisely the class of decomposable models (note that in non-chordal graphs, no matter how we direct the arrows, there will always be a pair of nonadjacent parents sharing a common child, a configuration that causes separation in undirected graphs but does not produce d-separation in dags).

Another crucial property of chordal graphs is that their *cliques* (i.e., the largest subgraphs whose nodes are all adjacent to each other) can be joined to form a tree $T$, called the *join tree*, such that any two cliques containing a node $\alpha$ are either adjacent in $T$ or connected by a chain of $T$ made entirely of cliques that contain $\alpha$ (Beeri et al., 1983) (an example is depicted in Figure 2).

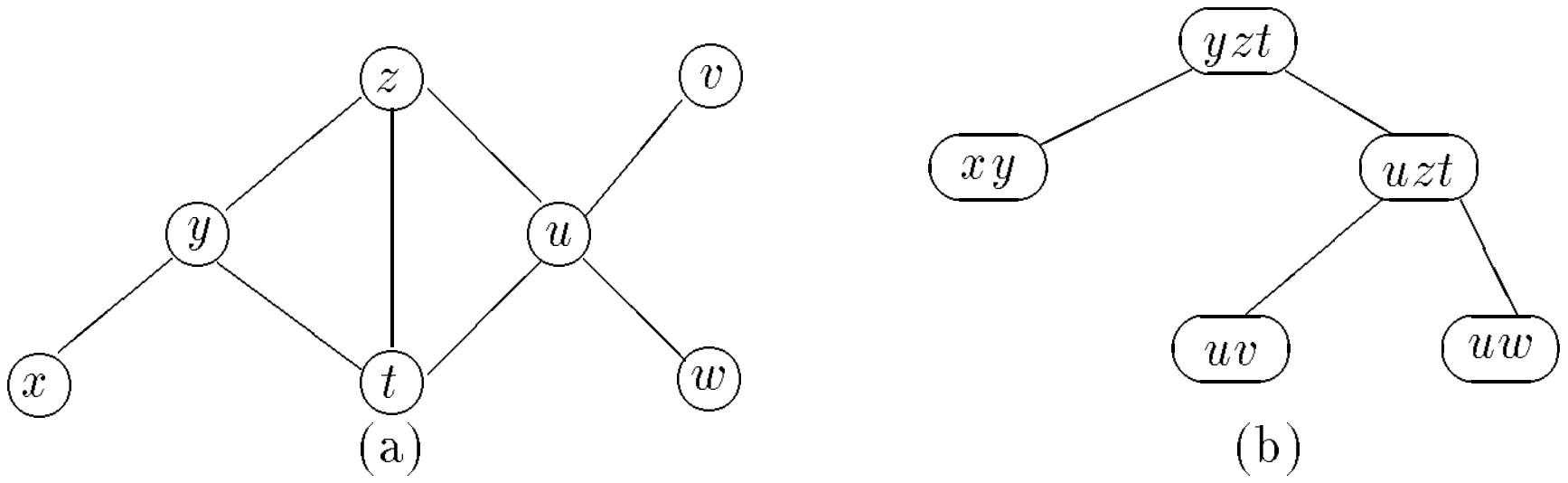

Figure 2: Chordal graph (a) and its join tree (b)

This result has important consequences for probabilistic modelling: the joint probability distribution factorises into the product of marginal distributions on cliques (Lauritzen et al., 1984; Pearl, 1988; Whittaker, 1991); moreover, maximum likelihood estimates of the model are directly calculable (Whittaker, 1991). As a consequence the compatibility functions used to quantitatively specify the model, have a clear meaning and can be easily estimated. Additionally, the tree structure of the cliques in a chordal graph facilitates recursive updating of probabilities. In fact, one of the most important algorithms for propagation (i.e., updating using local computations) of probabilities in dags, is based on a transformation of the given dag into a chordal graph, by moralising and next triangulating the dag (Lauritzen & Spiegelhalter, 1988).

## 4. Characterizing Decomposable Models

Our purpose is to find a characterization of decomposable models (or equivalently, of chordal graphs) in terms of properties of independence relationships. This will be carried out by adding a single property to the set of axioms, C1–C5, characterizing dependency models isomorphic to undirected graphs.

Let us consider the following axiom:

(C6) Strong Chordality:
$(I(\alpha,\beta|Z\cup\gamma\cup\delta)$ and $I(\gamma,\delta|U\setminus\{\gamma,\delta\}) \Rightarrow I(\alpha,\beta|Z\cup\gamma)$ or $I(\alpha,\beta|Z\cup\delta))\ \forall\alpha,\beta,\gamma,\delta\in U\ \forall Z\subseteq U\setminus\{\alpha,\beta,\gamma,\delta\}$.

This axiom establishes a condition that allows us to reduce the size of the conditioning set separating two variables $\alpha$ and $\beta$, namely that two of the variables in this set are conditionally independent. We are going to demonstrate that by adding the axiom of strong chordality to the axioms found by Pearl and Paz, C1–C5, the associated graph necessarily becomes a chordal graph and vice versa. Therefore, we shall obtain a characterization of decomposable models. Pearl (1988) proposed an axiom slightly different from C6, which is a necessary, though not sufficient condition for chordality. He called this axiom chordality:

(C7) Chordality:
$(I(\alpha,\beta|\gamma\cup\delta)$ and $I(\gamma,\delta|\alpha\cup\beta) \Rightarrow I(\alpha,\beta|\gamma)$ or $I(\alpha,\beta|\delta))\ \forall\alpha,\beta,\gamma,\delta\in U$.

Observe that in our context, i.e., assuming that C1–C5 hold, C6 implies C7: from $I(\alpha,\beta|\gamma\cup\delta)$ and $I(\gamma,\delta|\alpha\cup\beta)$, as strong union (C3) guarantees that $I(\gamma,\delta|U\setminus\{\gamma,\delta\})$ is implied by $I(\gamma,\delta|W)$ for any $W\subseteq U\setminus\{\gamma,\delta\}$ (in particular for $W=\{\alpha,\beta\}$), then by applying C6 with $Z=\emptyset$, we obtain $I(\alpha,\beta|\gamma)$ or $I(\alpha,\beta|\delta)$. However, the set of axioms C1–C5 and C7 do not constitute a characterization of chordal graphs, as the graph depicted in Figure 1 (b) shows: this graph is not chordal, but it satisfies C1–C5 and C7. By using C6 instead of C7 we shall obtain the desired result.

**Theorem 2** *A dependency model $M$ is isomorphic to a chordal graph if, and only if, it satisfies the axioms C1–C6.*

**Proof:** First, let us prove the sufficient condition. Using the Pearl and Paz result, from C1–C5 we deduce that $M$ is isomorphic to its associated graph $G$, and therefore independence in $M$ is equivalent to separation in $G$. We only have to prove that $G$ is chordal.

Let us suppose that $G$ is not chordal. Then, in $G$, there is a cycle $t_1t_2\ldots t_{n-1}t_nt_1$, $n\geq 4$, without a chord, i.e., $\forall i,j$ s.t. $1\leq i<i+1<j\leq n$, the edges $t_i$–$t_j$ do not belong to $E_M$ (except the edge $t_1$–$t_n$).

Let us consider the nodes $t_1$ and $t_{n-1}$, and the set of nodes $Z = U \setminus \{t_1, \ldots, t_n\}$. First, we are going to prove that the independence statement $I(t_1, t_{n-1}|Z \cup t_2 \cup t_n)$ has to be true: if it were $\neg I(t_1, t_{n-1}|Z \cup t_2 \cup t_n)$ then we could find a chain linking $t_1$ and $t_{n-1}$ not containing nodes from $Z \cup t_2 \cup t_n$, i.e., a chain linking $t_1$ and $t_{n-1}$ containing only nodes from $\{t_3, \ldots, t_{n-2}\}$; but in this case we would have an edge linking $t_1$ and some node $t_j$, $3 \leq j \leq n-2$, and this contradicts the assumption that the cycle has no chord. Therefore, we have $I(t_1, t_{n-1}|Z \cup t_2 \cup t_n)$.

On the other hand, the nodes $t_2$ and $t_n$ are not connected by any edge (once again because the cycle has not any chord), so they are separated by $U \setminus \{t_2, t_n\}$, and therefore we have $I(t_2, t_n|U \setminus \{t_2, t_n\})$. Now, using C6, we deduce either $I(t_1, t_{n-1}|Z \cup t_2)$ or $I(t_1, t_{n-1}|Z \cup t_n)$. In either case there is a chain linking $t_1$ and $t_{n-1}$ which is not blocked by the separating set: in the first case the chain is $t_1 t_n t_{n-1}$, and in the second case it is $t_1 t_2 \ldots t_{n-2} t_{n-1}$. Therefore, we obtain a contradiction, hence the graph $G$ has to be chordal.

Now, let us prove the necessary condition. Once again using Pearl and Paz's result, as $M$ is isomorphic to a graph $G$, then the properties C1–C5 hold.

Let us suppose that C6 does not hold. Then, we can find nodes $\alpha, \beta, \gamma, \delta$ and a subset of nodes $Z$ such that $I(\alpha, \beta|Z \cup \gamma \cup \delta)$, $I(\gamma, \delta|U \setminus \{\gamma, \delta\})$, $\neg I(\alpha, \beta|Z \cup \gamma)$ and $\neg I(\alpha, \beta|Z \cup \delta)$.

From $\neg I(\alpha, \beta|Z \cup \gamma)$ we deduce that a chain $\alpha t_1 \ldots t_n \beta$ exists in $G$, such that $t_i \notin Z \cup \gamma$ $\forall i$, i.e., $\{t_1 \ldots t_n\} \cap (Z \cup \gamma) = \emptyset$. However, from $I(\alpha, \beta|Z \cup \gamma \cup \delta)$ we know that every chain linking $\alpha$ and $\beta$ must contain some node from $Z \cup \gamma \cup \delta$. In particular, for the previously found chain, we have $\{t_1 \ldots t_n\} \cap (Z \cup \gamma \cup \delta) \neq \emptyset$. Therefore, there is a node $t_k$ such that $t_k = \delta$. Let us consider the node $t_{k-1}$: from $I(\alpha, \beta|Z \cup \gamma \cup \delta)$ and transitivity (C5), we obtain $I(\alpha, t_{k-1}|Z \cup \gamma \cup \delta)$ or $I(t_{k-1}, \beta|Z \cup \gamma \cup \delta)$. The first independence assertion cannot be true, because the chain $\alpha t_1 \ldots t_{k-2} t_{k-1}$ does not contain any node from $Z \cup \gamma \cup \delta$. Therefore, we have $I(t_{k-1}, \beta|Z \cup \gamma \cup \delta)$. If it were $I(t_{k-1}, \beta|Z \cup \gamma)$, then, from transitivity, we would obtain $I(\alpha, t_{k-1}|Z \cup \gamma)$ or $I(\alpha, \beta|Z \cup \gamma)$, and both statements are false, the first one because of the existence of the chain $\alpha t_1 \ldots t_{k-2} t_{k-1}$ and the second one because of the hypothesis. So, we have $\neg I(t_{k-1}, \beta|Z \cup \gamma)$. The same reasoning allows us to assert from $\neg I(\alpha, \beta|Z \cup \delta)$ that $\neg I(t_{k-1}, \beta|Z \cup \delta)$. So, we have found a node $t_{k-1}$ adjacent to $\delta = t_k$ satisfying the same properties as $\alpha$. A completely analogous reasoning applied to node $t_{k+1}$ proves $I(t_{k-1}, t_{k+1}|Z \cup \gamma \cup \delta)$, $\neg I(t_{k-1}, t_{k+1}|Z \cup \gamma)$, and $\neg I(t_{k-1}, t_{k+1}|Z \cup \delta)$. So, we have replaced nodes $\alpha$ and $\beta$ by two nodes adjacent to $\delta$ satisfying the same properties. Note that the case in which one or the other of $t_{k-1}$ and $t_{k+1}$ is $\alpha$ or $\beta$ does not matter to the subsequent argument.

Now, from $\neg I(t_{k-1}, t_{k+1}|Z \cup \delta)$ and $I(t_{k-1}, t_{k+1}|Z \cup \gamma \cup \delta)$ we deduce that there is a chain $t_{k-1} s_1 \ldots s_m t_{k+1}$ in $G$ such that $s_i \notin Z \cup \delta$ $\forall i$ and for some node $s_h$, $s_h = \gamma$. To simplify the notation, let us call $s_0 = t_{k-1}$, $s_{m+1} = t_{k+1}$. We can assume that $\forall i, j$ such that $0 < i + 1 < j \leq h$, there is no edge linking $s_i$ and $s_j$ (if this is not the case, we can simply replace the subchain $s_i s_{i+1} \ldots s_{j-1} s_j$ by the single edge $s_i$–$s_j$, i.e., we consider the shortest subchain between $t_{k-1}$ and $s_h$). For the same reason, we can also suppose that $\forall p, q$ such that $h < p + 1 < q \leq m + 1$, there is not any edge linking $s_p$ and $s_q$.

We have found a cycle $\delta s_0 s_1 \ldots s_{h-1} \gamma s_{h+1} \ldots s_m s_{m+1} \delta$ in $G$. Now, let $s_f$ and $s_g$ be two nodes satisfying $f < h < g$, $s_f$ and $s_g$ are adjacent to $\delta$ but $\delta$ is not adjacent to $s_j$ for all $j$ s.t. $f < j < g$ and $j \neq h$ (note that we can always find these two nodes, starting from $f = 0$ and $g = m + 1$). We still have a cycle $\delta s_f \ldots s_{h-1} \gamma s_{h+1} \ldots s_g \delta$ of length four or more, so that, according to the hypothesis, this cycle must have some chord. However, taking into account how the cycle has been constructed, the only possible chords are the edge $\gamma$–$\delta$ or an edge linking a node $s_i$, $f < i < h$, and a node $s_p$, $h < p < g$. The first possibility contradicts the hypothesis $I(\gamma,\delta|U\setminus\{\gamma,\delta\})$, and the second one implies the existence of the chain $t_{k-1}s_1\ldots s_f\ldots s_i s_p\ldots s_g\ldots s_m t_{k+1}$ linking $t_{k-1}$ and $t_{k+1}$, which does not contain any node from $Z\cup\gamma\cup\delta$, in contradiction with the statement $I(t_{k-1},t_{k+1}|Z\cup\gamma\cup\delta)$. Therefore, the property C6 has to be true.

□

We can establish another interesting characterization of chordal graphs, by also adding only one axiom to those of Pearl and Paz. This new axiom is the following:

(C8) Clique-separability:
$(I(\alpha,\beta|U\setminus\{\alpha,\beta\})\Rightarrow \exists W\subseteq U\setminus\{\alpha,\beta\}$ such that $I(\alpha,\beta|W)$ and either $|W|\leq 1$ or
$\neg I(\gamma,\delta|U\setminus\{\gamma,\delta\})\,\forall\gamma,\delta\in W)\ \forall\alpha,\beta\in U$.

Axiom C8 asserts that whenever two nodes $\alpha$ and $\beta$ are not adjacent (are independent), we can find a separating set whose nodes are all adjacent to each other, i.e., a complete separating set.

**Theorem 3** *A dependency model $M$ is isomorphic to a chordal graph if, and only if, it satisfies the axioms C1–C5 and C8.*

**Proof:** Let us prove the necessary condition. As the graph $G$ associated to $M$ is chordal, from Theorem 2 we know that the properties C1–C5 and C6 hold.

Let us suppose that C8 does not hold. Then, there are $\alpha$ and $\beta$, such that $I(\alpha,\beta|U\setminus\{\alpha,\beta\})$ but for all $W\subseteq U\setminus\{\alpha,\beta\}$ it is either $\neg I(\alpha,\beta|W)$, or $|W|>1$ and $\exists\gamma,\delta\in W$ such that $I(\gamma,\delta|U\setminus\{\gamma,\delta\})$.

Let $W_0$ be any separating set of minimal size for $\alpha$ and $\beta$, i.e., $I(\alpha,\beta|W_0)$ and $\neg I(\alpha,\beta|S)$ $\forall S\subset W_0$ (we know that at least one separating set of this type has to exist, because $I(\alpha,\beta|U\setminus\{\alpha,\beta\})$ holds). Then, we can deduce that $|W_0|>1$ and $\exists\gamma,\delta\in W_0$ such that $I(\gamma,\delta|U\setminus\{\gamma,\delta\})$. Let us define $Z=W_0\setminus\{\gamma,\delta\}$. Thus, we have $I(\alpha,\beta|Z\cup\gamma\cup\delta)$ and $I(\gamma,\delta|U\setminus\{\gamma,\delta\})$ and, by applying C6, we obtain either $I(\alpha,\beta|Z\cup\gamma)$ or $I(\alpha,\beta|Z\cup\delta)$, i.e., $I(\alpha,\beta|W_0\setminus\{\delta\})$ or $I(\alpha,\beta|W_0\setminus\{\gamma\})$, which contradicts the minimality of $W_0$. Therefore, C8 has to be true.

To prove the sufficient condition, let us suppose that $G$ is not chordal. Then, there is a cycle $t_1t_2\ldots t_{n-1}t_nt_1$, $n\geq 4$, without a chord. So, the nodes $t_1$ and $t_{n-1}$ are not adjacent, hence they are separated, and let $W$ be any separating set of $t_1$ and $t_{n-1}$, i.e., satisfying $I(t_1,t_{n-1}|W)$. Then $t_n\in W$ and $\{t_2,\ldots,t_{n-2}\}\cap W\neq\emptyset$, otherwise we could find a chain linking $t_1$ and $t_{n-1}$ which would not be blocked by $W$, thus contradicting $I(t_1,t_{n-1}|W)$. So, every separating set $W$ contains $t_n$ and some $t_i, 2\leq i\leq n-2$, hence $|W|>1$. Now, by applying C8, we deduce that $\neg I(t_n,t_i|U\setminus\{t_n,t_i\})$, i.e., $t_n$ and $t_i$ are adjacent nodes, which contradicts the assumption that the cycle had no chord. Then, the conclusion is that the graph has to be chordal.

□

## 5. Relationships with other Characterizations of Decomposable Models

There is a characterization of decomposable models[1] (Lauritzen, 1989) which is quite related to ours: *an undirected graph is chordal if, and only if, every subset of nodes that separates any two nodes $\alpha$ and $\beta$ and is minimal is complete.*

In order to rewrite this result using our notation, let us consider the following axiom:

(C9) Completeness:
$(I(\alpha,\beta|Z)$ and $\neg I(\alpha,\beta|S) \forall S \subset Z \Rightarrow |Z| \leq 1$ or $\neg I(\gamma,\delta|U \setminus \{\gamma,\delta\})\ \forall \gamma,\delta \in Z)\ \forall \alpha,\beta \in U\ \forall Z \subseteq U \setminus \{\alpha,\beta\}$.

Axiom C9 says exactly that any minimal separator of $\alpha$ and $\beta$ has to be complete. An equivalent formulation of this axiom reads: any separator of $\alpha$ and $\beta$ which is not complete cannot be minimal. In symbols:

(C9') Completeness:
$(I(\alpha,\beta|Z \cup \gamma \cup \delta)$ and $I(\gamma,\delta|U \setminus \{\gamma,\delta\}) \Rightarrow \exists W \subset Z \cup \gamma \cup \delta$ such that $I(\alpha,\beta|W))\ \forall \alpha,\beta,\gamma,\delta \in U\ \forall Z \subseteq U \setminus \{\alpha,\beta,\gamma,\delta\}$.

Then, Lauritzen's result can be reformulated as follows: *A dependency model $M$ is isomorphic to a chordal graph if, and only if, it satisfies the axioms C1–C5 and either C9 or C9'.*

Note the similarity between C9 and C8 and between C9' and C6. Taking into account Theorems 2 and 3, we can deduce that all these axioms, C6, C8, C9 and C9', are equivalent among each other (assuming that C1–C5 hold). However, this equivalence is not evident, in spite of the similarities among axioms: it is clear that C6 implies C9' and C9 implies C8, but the opposite implications are not obvious. In fact, strong chordality and clique-separability seem stronger and weaker, respectively, than completeness. This becomes clearer if we express the axioms in the following way: Assuming that two nodes $\alpha$ and $\beta$ can be separated:

- *Completeness (C9 or C9'):* If a separator of $\alpha$ and $\beta$ is minimal, then it is complete; or, equivalently, if a separator of $\alpha$ and $\beta$ is not complete, then it has a proper subset which is still a separator of $\alpha$ and $\beta$.

- *Clique-separability (C8):* There exists a separator of $\alpha$ and $\beta$ which is complete.

- *Strong chordality (C6):* If a separator of $\alpha$ and $\beta$ is not complete, then it has a proper subset which is still a separator of $\alpha$ and $\beta$; moreover, we can find this subset by removing, from the initial separator, one of the nodes causing its incompleteness.

Observe that both C6 and C9' share the same antecedent, but the consequent of C9' only says that there exists a separator, whereas the consequent of C6 gives more information about the identity of this separator. Note also that both C8 and C9 assert the existence of a complete separator, but C9 requires a previous condition (minimality) and C8 does not.

1. The existence of this result was pointed out to me by a reviewer.

## 6. Concluding Remarks

We have found two new characterizations of the class of decomposable dependency models, in terms of properties of independence relationships. We believe that these results are theoretically interesting, because they provide a new perspective of this important and well studied class of graphical models. Moreover, our results are quite concise, since only one property has to be added to the set of properties characterizing independence relationships in undirected graphs. They could also be useful for proving results about models of this sort.

From a more practical point of view, the axiomatic characterizations create desiderata that could drive automated construction of chordal graphs from data. As we have already commented, practical use of graphical models and, particularly, of bayesian networks, requires that the dag representing the model be transformed into a chordal graph. From the perspective of learning models from data, it may be interesting to estimate directly the chordal graph from the available data, instead of first learning the dag and after converting it into a chordal graph. We believe that the basic independence properties of chordal graphs identified by our theoretical study, C6 and C8, could guide us in the design of efficient algorithms for learning chordal graphs. It is known that the problem of learning bayesian networks from data is computationally very complex. For example, some algorithms (Spirtes, Glymour, & Scheines, 1993) start from a complete undirected graph, and then try to remove edges by testing for conditional independence between the linked nodes, but using conditioning sets as small as possible (thus reducing the complexity and increasing reliability). In this context, if we rewrite the property C6 in the following way:

$$\neg I(\alpha,\beta|Z\cup\gamma) \text{ and } \neg I(\alpha,\beta|Z\cup\delta) \text{ and } I(\alpha,\beta|Z\cup\gamma\cup\delta) \;\Rightarrow\; \neg I(\gamma,\delta|U\setminus\{\gamma,\delta\}),$$

then we could use it as a rule that simultaneously allows us to remove the edge $\alpha$–$\beta$ from the current graph, and to fix the edge $\gamma$–$\delta$ as a true edge in the graph.

Similarly, the property C8 could give rise to the following rule: if we are trying to remove an edge $\alpha$–$\beta$ from the current graph, by testing conditional independence statements like $I(\alpha,\beta|W)$, then discard as candidate separating sets those sets $W$ whose nodes are not all adjacent to each other.

This topic of designing efficient algorithms for learning chordal graphs will be the object of future research.

## Acknowledgements

This work has been supported by the Spanish Comisión Interministerial de Ciencia y Tecnología (CICYT) under Project n. TIC96-0781. I would like to thank Milan Studený and three anonymous reviewers for helpful comments and suggestions. I am particularly grateful to the reviewer who pointed out to me the existence of Lauritzen's characterization of chordal graphs.